\documentclass[english,10pt, a4paper, twocolumn,utf8]{article}
\usepackage[T1]{fontenc}
\usepackage[utf8]{inputenc}
\usepackage{babel}
\usepackage{textcomp}
\usepackage{graphicx}
\usepackage[unicode=true,
 bookmarks=true,bookmarksnumbered=true,bookmarksopen=false,
 breaklinks=false,pdfborder={0 0 1},backref=false,colorlinks=false]
 {hyperref}
\hypersetup{pdftitle={Data Augmentation of Spectral Data for ConvolutionalNeural Network (CNN) Based Deep Chemometrics},
 pdfauthor={Esben Jannik Bjerrum, Mads Glahder and Thomas Skov},
 pdfsubject={Deep Chemometrics},
 pdfkeywords={Deep Chemometrics, NIR, CNN, PLS}}

\makeatletter

\providecommand{\tabularnewline}{\\}

\usepackage[]{babel} 

\usepackage{amsmath,amsfonts,amsthm} 

\usepackage[svgnames]{xcolor} 

\usepackage[small, labelfont=bf, up, textfont=it]{caption} 

\usepackage{booktabs} 

\usepackage{lastpage} 

\usepackage{graphicx} 

\usepackage{enumitem} 
\setlist{noitemsep} 

\usepackage{sectsty} 
\allsectionsfont{\usefont{OT1}{phv}{b}{n}} 

\usepackage{mdframed} 

\usepackage[font=small,labelfont=bf]{caption}
\makeatletter
\g@addto@macro\@floatboxreset\centering
\makeatother

\usepackage{geometry} 

\usepackage[T1]{fontenc} 
\usepackage[utf8]{inputenc} 

\usepackage{XCharter} 

\usepackage{fancyhdr} 
\newcommand{\headrulecolor}[1]{\patchcmd{\headrule}{\hrule}{\color{#1}\hrule}{}{}}
\newcommand{\footrulecolor}[1]{\patchcmd{\footrule}{\hrule}{\color{#1}\hrule}{}{}}

\renewcommand{\footrulewidth}{1pt} 

\headrulecolor{DarkGoldenrod}
\footrulecolor{DarkGoldenrod}
\fancypagestyle{firstpage}{ 
\renewcommand{\footrulewidth}{0.0pt} 
}

\newcommand{\authorstyle}[1]{{\large\usefont{OT1}{phv}{b}{n}\color{DarkRed}#1}} 

\newcommand{\institution}[1]{{\footnotesize\usefont{OT1}{phv}{m}{sl}\color{Black}#1}} 

\usepackage{titling} 

\pretitle{
\date{}
\vspace{-70pt} 
\vspace{35pt} 
\fontsize{18}{24}\usefont{OT1}{phv}{b}{n}\selectfont 
\color{DarkRed} 
}

\posttitle{\par\vskip 15pt} 

\preauthor{} 

\postauthor{
\vspace{-40pt} 
}

\usepackage{lettrine} 
\usepackage{fix-cm}	

\newcommand{\initial}[1]{ 
\lettrine[lines=3,findent=4pt,nindent=0pt]{
\color{DarkGoldenrod}
{#1}
}{}%
}

\usepackage{xstring} 

\newcommand{\lettrineabstract}[1]{

\mdframed[backgroundcolor=gray!20,hidealllines=true]
\vspace{5pt} 
\StrLeft{#1}{1}[\firstletter] 
\initial{\firstletter}\textbf{\StrGobbleLeft{#1}{1}} 
\vspace{5pt} 
\endmdframed 

\vspace{10pt} 
}

\author{
\authorstyle{Esben Jannik Bjerrum\textsuperscript{1,*}, Mads Glahder\textsuperscript{1} and Thomas Skov\textsuperscript{2}} 
\newline\newline 
\textsuperscript{1}\institution{Wildcard Pharmaceutical Consulting, Zeaborg Science Center, Frødings Allé 41, 2860 Søborg, Denmark.}\\ 
\textsuperscript{2}\institution{Spectroscopy and Chemometrics, Department of Food Science, Copenhagen University, Rolighedsvej 26, 1958 Frederiksberg, Denmark}\\ 
\textsuperscript{*}\institution{Corresponding Author: \href{mailto://esben@wildcardconsulting.dk}{esben@wildcardconsulting.dk}} 
}

\makeatother

\begin{document}

\twocolumn[   \begin{@twocolumnfalse}

\title{Data Augmentation of Spectral Data for Convolutional Neural Network
(CNN) Based Deep Chemometrics}

\maketitle
\thispagestyle{fancy}
\renewcommand{\footrulewidth}{0.0pt}
\lhead{}
\chead{}
\rhead{}
\lettrineabstract{Deep learning methods are used on spectroscopic data to predict drug content in tablets from near infrared (NIR) spectra. Using convolutional neural networks (CNNs), features are extracted from the spectroscopic data. Extended multiplicative scatter correction (EMSC) and a novel spectral data augmentation method are benchmarked as preprocessing steps. The learned models perform better or on par with hypothetical optimal partial least squares (PLS) models for all combinations of preprocessing. Data augmentation with subsequent EMSC in combination gave the best results. The deep learning model CNNs also outperform the PLS models in an extrapolation challenge created using data from a second instrument and from an analyte concentration not covered by the training data. Qualitative investigations of the CNNs kernel activations show their resemblance to wellknown data processing methods such as smoothing, slope/derivative, thresholds and spectral region selection.}

 \end{@twocolumnfalse} ]

\section*{Introduction}

Use of artificial neural networks (ANNs) for analysis of spectroscopic
data is decades old\cite{Long1990} and they have been widely used
as nonlinear models for the analysis of spectroscopic data. Convolutional
neural networks (CNNs)\cite{Krizhevsky2012,LeCun1995} are ANNs with
a special layout and restrictions that make them excellent for modelling
spatial data. Neighboring data points are analysed in mini networks
by ``scanning'' the data with a limited spatial width (filter size).
This is done multiple times with different mini networks (kernels).
The trained weights for the mini network for each kernel must be the
same for the entire sweep of the data. In image analysis and classification
these ANN types surpassed human interpretation two years ago\cite{He2015}.
The activations of the kernels are then fed forward in the neural
network into new convolutional layers or standard dense layers.
\begin{figure}
\caption{\label{fig:DA_illustraion}Components of the spectral data augmentation.
Data augmentation is created with randomly scaled contributions from
offset, slope and multiplication to simulate baseline offset, slope
differences and differences in intensity in the spectral recordings.}

\includegraphics[width=1\columnwidth]{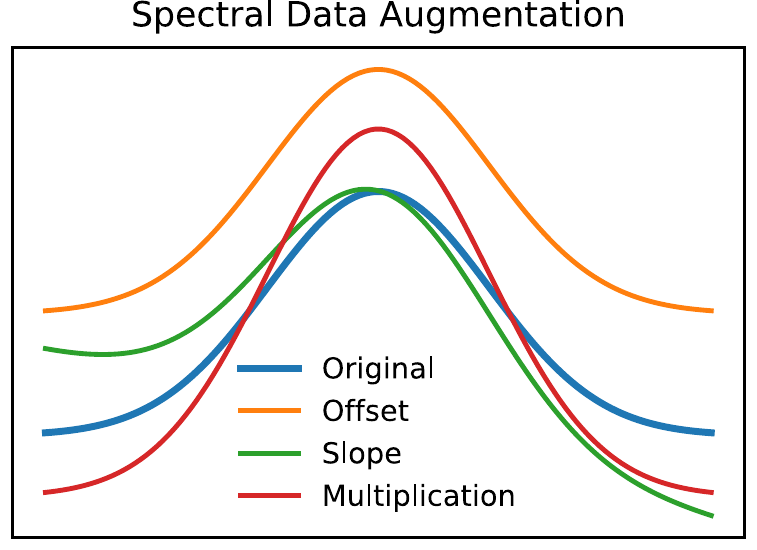}
\end{figure}

A spectrum can be thought of as a one-dimensional picture, and this
type of neural network architecture has recently been used for spectroscopical
data. As examples they were employed for classification of Raman Spectra\cite{Liu2017}
and for classification of pharmaceutical tablets using VIS-NIR spectroscopy\cite{Baik2017}.
CNNs have also been used for regression modelling of NIR and Raman
Data, including advanced spectral region selection via a custom regularization
function.\cite{Acquarelli2017}

Data augmentation is a wellknown technique for improving robustness
and training of neural networks\cite{Krizhevsky2012}. It has been
used with success in many domains ranging from image classification\cite{Wang}
to molecular modelling\cite{Bjerrum2017}. The core idea is to expand
the number of training samples from the limited number of labelled
samples by simulating various expected variations in the datasets.
For images it is easy to understand that a picture of a cat is recognisable
even though it has been rotated randomly between -45\textdegree{}
to 45\textdegree{} or flipped. The neural network subsequently learns
how to cope with such variations resulting in a more robust training.
For spectral data the variation employed is suggested as in Figure
\ref{fig:DA_illustraion}. Here random offsets, random changes in
slope and random multiplications are added to the existing spectrums
to expand the dataset.

Dropout\cite{Srivastava2014} is a wellknown regularization technique
employed for deep learning neural network models. During training
each activation (or data input point) is randomly dropped/removed
from the vector fed into the subsequent layer with a tunable probability
(the dropout rate). The effect is a disruption of the neuron interdependence,
so that each neuron can't make predictions on very specific patterns
of the input. This in turn leads to more independent neurons and redundancy
in the data handling. This limits the overfitting capacaity and give
better generalization for prediction of novel samples. After training
the weights of the layer are scaled proportionally to the dropout
rate and the dropout layer is not active during sampling.

Here we analyze a tablet dataset\cite{ShootoutWayback,EigenVectorDataset}
using CNNs with automatic tuning of the model hyperparameters and
regularization level in the form of dropout layers. The dataset consists
of assay results from analysis of pharmaceutical tablets and NIR spectra
recorded with two different instruments. Data preprocessing in the
form of spectral data augmentation is implemented and the perfomance
compared with extended multiplicative scatter correction (EMSC). All
dataset treatments are compared to hypothetical optimal PLS models
as a baseline method. The model types performances are also compared
on a specially crafted extrapolation challenge for both assay result
values and instrument recordings.

\section*{Methods}

\subsubsection*{Datasets}

The dataset was downloaded\cite{EigenVectorDataset} and read from
Matlab format into Python using utilities from Scientific Python\cite{Scipy2001}.
The original data from instrument one and two was pooled on their
own, and divided into different training, validation and external
test sets. A standard dataset split was created by randomly splitting
off 20\% for test and validation set in successive steps. The extrapolation
set was split into sets with the test set having measured concentration
above 228mg, the validation set between 228mg and 212mg and the training
set with concentrations below 212mg. The training and validation sets
was exclusively done with spectrums obtained with instrument one,
whereas the test sets were exclusively taken from instrument two.
The spectral region from 600 to 1798 nm was used for modelling.

\paragraph*{Outlier removal}

Spectral outliers were identified by PLS modeling, using the implementation
of the NIPALS algorithm\cite{Geladi1986} in Scikit-Learn\cite{Pedregosa2011},
without scaling and a maximum of 100.000 iterations and a tolerance
of 10\textsuperscript{-16}.The entire dataset was subjected to 10
fold cross validation (CV) varying the number of principal components
between 1 and 30. The number of principal components that gave the
lowest average Huber loss\cite{Huber1964} for the cross validation
sets was used to model the entire dataset and the outliers identified
as samples having more than 2.5 times the standard deviation of the
absolute error of prediction.

\paragraph*{Data Augmentation}

Some of the datasets were augmented by adding random variations in
offset, multiplication and slope. Offset was varied $\pm$0.10 times
the standard deviation of the training set. Multiplication was done
with 1$\pm$0.10 times the standard deviation of the training set,
and the slope was adjusted uniformly randomly between 0.95 to 1.05.
This was done 9 times for each sample in the training set and appended
to the dataset. The assay reference set was expanded appropriately.

\paragraph*{Extended Multiplicative Scatter Correction}

Some of the datasets were subjected to Extended Multiplicative Scatter
Correction (EMSC)\cite{Martens1991} with an order of one. The EMSC
python code was adapted from the ChemPy project\cite{Jarvis2006}.
The reference spectrum was the average of the training set for EMSC
correction of all the dataset subsets.

\paragraph*{Normalization}

All datasets were normalized by subtracting the global mean of the
training set and dividing by two times the global standard deviation
of the training ensuring that most of the values are in the range
-1 to 1.

\subsubsection*{Neural Networks}

A CNN was built by using Keras v.1.2.2\cite{chollet2015} with Theano
v.0.9.0\cite{2016arXiv160502688full} as the computation back end.
GPU computation was done using CUDA version 8.0.\cite{Nickolls2008}
The input was fed to a Gaussian noise layer with a standard deviation
of 0.01, followed by two one-dimensional convolutional layers\cite{Zeiler2010,Dumoulin2016}
with a rectified linear activation.\cite{Nair2010} The output from
the convolutional kernels were flattened and a dropout layer\cite{Srivastava2014}
was added before connecting to a fully connected dense layer with
a linear activation function. The output layer was a single dense
neuron with a linear activation function. The loss function was a
Huber loss\cite{Huber1964} adapted for use with Keras. Training was
done with the Adadelta optimizer\cite{Zeiler2012}.

\subsubsection*{Hyperparameter tuning and training}

The hyperparameters for the neural network listed in Table \ref{tab:Hyper-parameter-Search}
\begin{table}
\caption{\label{tab:Hyper-parameter-Search}hyperparameter search space, l1:
Layer 1, l2: Layer 2, Conv: Convolutional}

\centering{}%
\begin{tabular}{lcc}
\hline 
\multicolumn{1}{c}{Parameter} & Search Space & Type\tabularnewline
\hline 
Conv l1 no. Kernels & 2 - 40 & Integer\tabularnewline
Conv l1 Filter size & 5 - 150 & Integer\tabularnewline
Conv l2 no. Kernels & 2 - 40 & Integer\tabularnewline
Conv l2 Filter size & 5 - 150 & Integer\tabularnewline
Dropout after Flatten & 0 - 0.5 & Float\tabularnewline
Dense no. Neurons & 4 - 1000 & Integer\tabularnewline
\hline 
\end{tabular}
\end{table}
 was optimized using Bayesian optimization with Gaussian processes
as implemented in the Python package GpyOpt\cite{gpyopt2016} version
1.0.3. The optimization was initialized with 20 random parameter sets
followed by up to 40 iterations of standard Gaussian process optimization
with the expected improvement acquisition function. In each iteration
a new network was built from the parameter set and trained for 40
or 200 epochs, with learning rates of 0.084 and 0.094 for datasets
with and without data augmentation, respectively. Batch size was set
to 45. The loss function was taken as an average of the validation
loss observed for the last 10 epochs of training. 

Final training of the neural networks was done with the identified
optimized architecture and hyperparameters for each dataset and preprocessing
type. The training was extended to 100 and 250 epochs for datasets
with and without data augmentation, respectively. The learning rate
was reduced with a factor of two when the validation loss failed to
improve for 10 or 25 epochs (DA/non-DA).

The number of components for the PLS model was tuned by identifying
the number of components that gave the lowest Huber loss on the test
set when trained on the pooled validation and training sets. Additionally
CV loss and training loss were recorded and used to calculate a corrected
CV loss, where the difference between the training loss and the CV
loss is added to the CV loss.

All computations and training were done on a Linux workstation (Ubuntu
Mate 16.04 LTS) with 32GB of RAM, i5-2405S CPU @ 2.50GHz and an Nvidia
Geforce GTX1060 graphics card with 6 GB of RAM.

\section*{Results}

\begin{figure}
\caption{\label{fig:Opt_convergense}Example convergence of the hyperparameter
tuning (EMSC dataset). The blue line shows the loss for the validation
set at a given iteration. The first 20 iterations are randomly selected
hyperparameters and the rest are hyperparameters chosen by the bayesian
optimizer from GPyOpt. The dashed grey line shows the loss of the
best found hyperparameter set at a given iteration. The best solution
was found at iteration 41.}

\includegraphics[width=1\columnwidth]{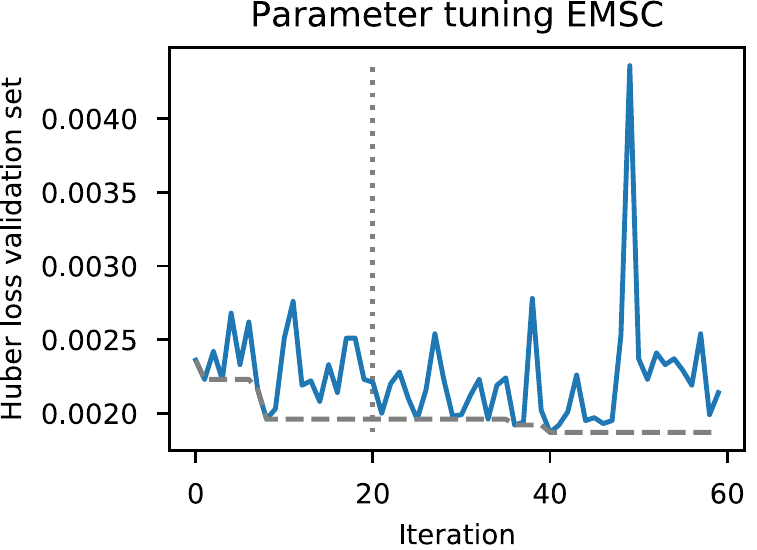}
\end{figure}

The two different ways of splitting into subsets and the different
selected preprocessing steps resulted in the datasets listed in Table
\ref{tab:Datasets}.
\begin{table*}
\caption{\label{tab:Datasets}Details of the datasets used. GS: Global scaling,
DA: Data Augmentation, EMSC: Extended multiplicative scatter correction}

\begin{tabular}{llccc}
\hline 
 &  &  & Size & \tabularnewline
\cline{3-5} 
\multicolumn{1}{c}{Dataset} & \multicolumn{1}{c}{Preprocessing} & Training & Validation & Test\tabularnewline
\hline 
Standard & GS & 404 & 101 & 136\tabularnewline
EMSC & EMSC + GS & 404 & 101 & 136\tabularnewline
Extrapolation (Ext.) & GS & 576 & 52 & 13\tabularnewline
Ext. EMSC & EMSC + GS & 576 & 52 & 13\tabularnewline
Data Augmentation (DA.) & DA + GS & 4040 & 1010 & 136\tabularnewline
DA EMSC & DA + EMSC + GS & 4040 & 1010 & 136\tabularnewline
Ext. DA & DA + GS & 5760 & 520 & 13\tabularnewline
Ext. DA EMSC & DA + EMSC + GS & 5760 & 520 & 13\tabularnewline
\hline 
\end{tabular}
\end{table*}
 The employed data augmentation resulted in a 10 fold increase in
training and validation set size, whereas the test set was not data
augmented. Splitting for the extrapolation datasets based on assay
values resulted in validation and test set sizes of 52 and 13, respectively.
Each datasets hyperparameters were optimized on its own, by optimizing
the predictive performance on the validation set for the CNN models.
An example of the convergence of the optimization is shown in Figure
\ref{fig:Opt_convergense}.
 After the first 20 randomly selected parameter sets, the optimization
algorithm chooses the next parameter set to test, finding the best
solution at iteration 41.

\begin{figure}
\caption{\label{fig:PLS_tuning}Example of the PLS hyperparameter tuning. The
performance using the Huber loss is followed as the number of principal
components is varied.}

\includegraphics[width=1\columnwidth]{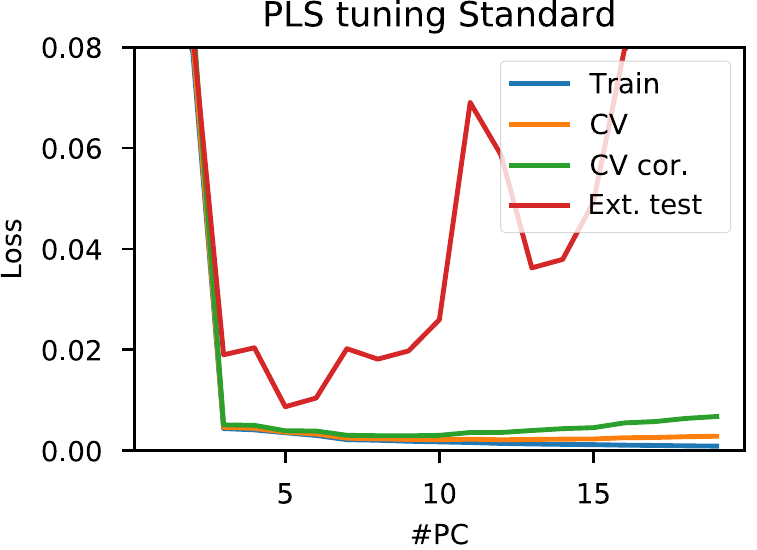}
\end{figure}

In contrast to the use of validation sets for tuning the CNNs, the
hyperparameters of the PLS models were optimised with respect to the
test set. An example of the tuning curve for the Standard set is shown
in Figure \ref{fig:PLS_tuning}.
 The red curve has the lowest loss at five components, whereas the
CV and Corrected CV have minimums at 10 and 9 components, respectively.
Using the minimum of the corrected validation loss or validation loss
directly in general resulted in larger number of PLS components to
be included and lower predictive performance for the test set. 

\begin{figure*}
\caption{\label{fig:Example-Training-History}Example training history comparing
a dataset with and without data augmentation.}

\includegraphics[width=1\textwidth]{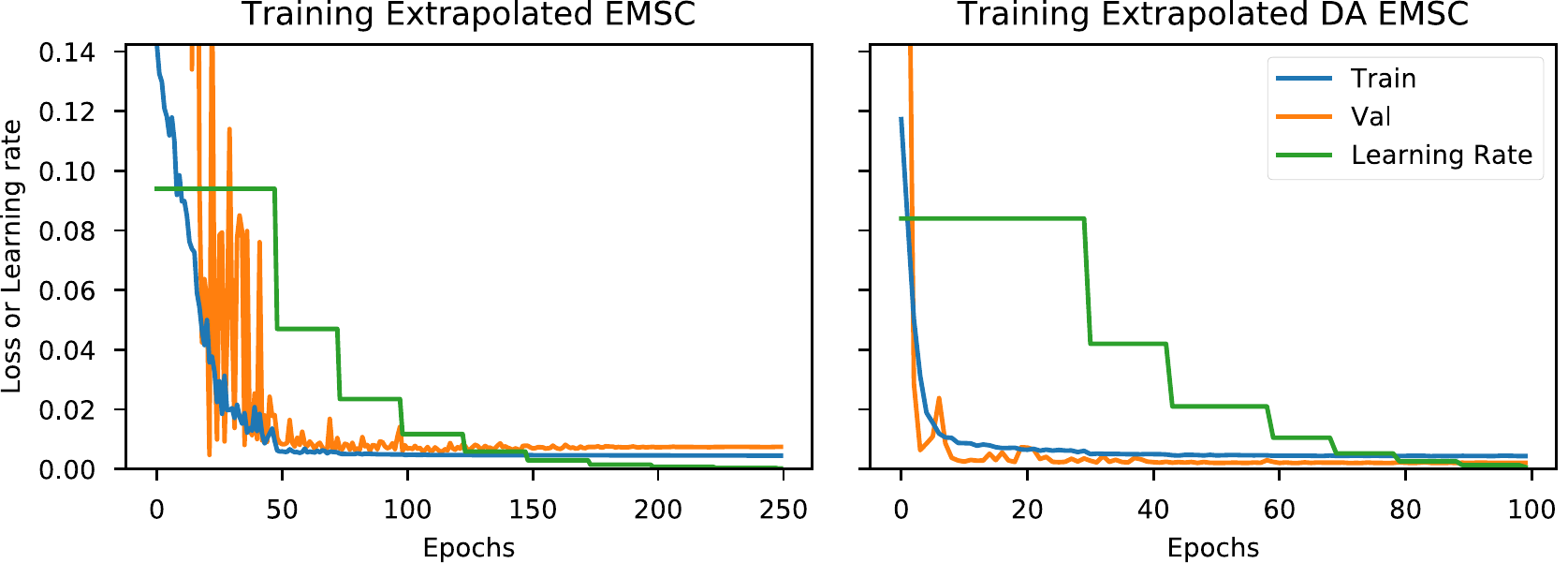}
\end{figure*}

The found hyperparameters are listed in Table \ref{tab:Optimized-Hyperparameters-wo DA}
\begin{table*}
\caption{\label{tab:Optimized-Hyperparameters-wo DA}Optimized hyperparameters
for datasets without data augmentation. Ext: Extrapolation, EMSC:
Extended multiplicative scatter correction}

\begin{tabular}{lcccc}
\hline 
\multicolumn{1}{c}{Parameter} & Standard & EMSC & Ext. & Ext. EMSC\tabularnewline
\hline 
Conv layer 1 no. Kernels & 14 & 18 & 22 & 40\tabularnewline
Conv layer 1 Filter size & 29 & 40 & 45 & 96\tabularnewline
Conv layer 2 no. Kernels & 30 & 18 & 25 & 40\tabularnewline
Conv layer 2 Filter size & 22 & 29 & 40 & 45\tabularnewline
Dropout after Flatten & 0.045 & 0.070 & 0.082 & 0.0\tabularnewline
Dense Layer no. Neurons & 176 & 266 & 468 & 470\tabularnewline
\hline 
Principal Components & 5 & 4 & 8 & 6\tabularnewline
\hline 
\end{tabular}
\end{table*}
 and \ref{tab:Optimized-Hyperparameters-with DA}
\begin{table*}
\caption{\label{tab:Optimized-Hyperparameters-with DA}Optimized hyperparameters
for datasets with data augmentation. Ext: Extrapolation, EMSC: Extended
multiplicative scatter correction.}

\begin{tabular}{lcccc}
\hline 
\multicolumn{1}{c}{Parameter} & DA & DA EMSC & Ext. DA & Ext. DA EMSC\tabularnewline
\hline 
Conv layer 1 no. Kernels & 29 & 19 & 7 & 22\tabularnewline
Conv layer 1 Filter size & 124 & 138 & 80 & 54\tabularnewline
Conv layer 2 no. Kernels & 28 & 5 & 24 & 29\tabularnewline
Conv layer 2 Filter size & 119 & 54 & 60 & 42\tabularnewline
Dropout after Flatten & 0.205 & 0.089 & 0.015 & 0.388\tabularnewline
Dense Layer no. Neurons & 289 & 527 & 749 & 289\tabularnewline
\hline 
Principal Components & 5 & 4 & 9 & 7\tabularnewline
\hline 
\end{tabular}
\end{table*}
 for the datasets without and with data augmentation, respectively.
The dropout rate is in the low range for most of the datasets. The
number of kernels for layer two seem on average to be larger or the
same as layer one with the DA EMSC dataset as an exception. Figure
\ref{fig:Example-Training-History}
 shows a comparison between the training histories for the EMSC dataset
with and without data augmentation. Both training histories in the
end obtain a low difference between the training and validation loss
for many epochs in the end of training, indicating absence of overfitting.
The validation loss displays a much rougher curve over the course
of training for the dataset without data augmentation.

\begin{figure*}
\caption{\label{fig:Scatter-plot}Scatter plot of true and predicted reference
values using the extrapolation dataset with data augmentation followed
by extended multiplicative scatter correction (EMSC). The training
of the CNN model included the validation set used in the tuning of
the hyperparameters. The small insert on each graph shows a zoom of
the external test set.}

\includegraphics[width=1\textwidth]{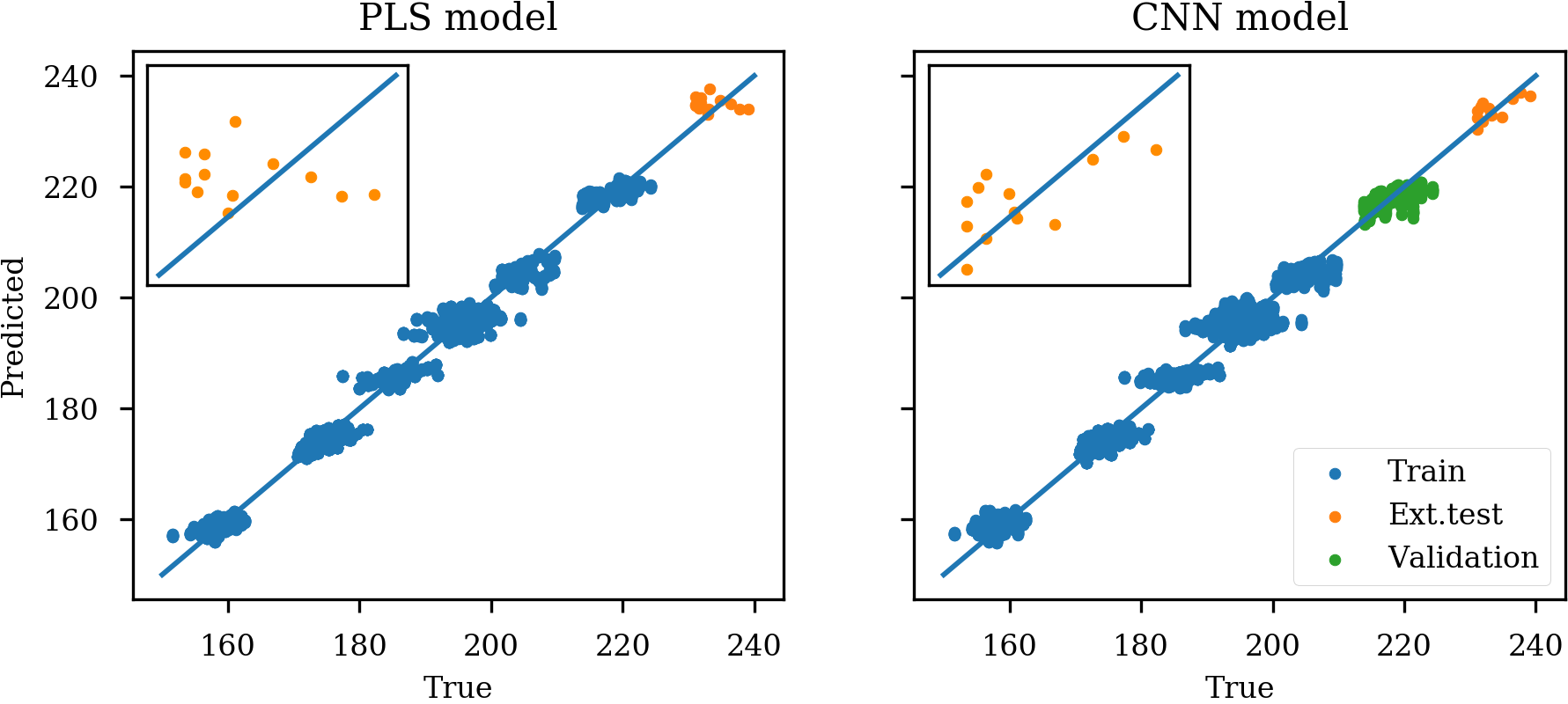}
\end{figure*}

The measured model performance for the different dataset splits and
preprocessing selections is shown in Table\ref{tab:Model-performance}.
\begin{table*}
\caption{\label{tab:Model-performance}Model performance on train and test
set for partial least squares (PLS) and neural network models (NN).
Train set includes the validation set used for tuning of hyperparameters
and consists of spectrum obtained with instrument one. Test set consists
of tablet samples that was never included in train and validation
sets and consists of spectrums obtained with instrument two. The best
performance on each test set is marked in bold.}

\begin{tabular}{lcccccccc}
\hline 
 &  &  & PLS &  &  &  & NN & \tabularnewline
\cline{3-5} \cline{7-9} 
\multicolumn{1}{c}{Dataset} & Subset & R\textsuperscript{2} & RMSE & Huber &  & R\textsuperscript{2} & RMSE & Huber\tabularnewline
\hline 
Standard & Train & 0.97 & 2.97 & 1.65 &  & 0.97 & 3.02 & 1.72\tabularnewline
 & Test & 0.94 & 4.43 & 2.60 &  & \textbf{0.97} & \textbf{4.01} & \textbf{2.51}\tabularnewline
\cline{2-9} 
EMSC & Train & 0.97 & 2.88 & 1.62 &  & 0.98 & 2.37 & 1.21\tabularnewline
 & Test & 0.96 & 4.15 & 2.71 &  & \textbf{0.97} & \textbf{3.17} & \textbf{1.91}\tabularnewline
\cline{2-9} 
Ext. & Train & 0.98 & 2.10 & 1.01 &  & 0.98 & 2.47 & 1.30\tabularnewline
 & Test & \textbf{0.36} & 8.03 & 6.29 &  & 0.35 & \textbf{4.59} & \textbf{3.22}\tabularnewline
\cline{2-9} 
Ext. EMSC & Train & 0.98 & 2.15 & 1.04 &  & 0.97 & 2.66 & 1.41\tabularnewline
 & Test & 0.18 & \textbf{3.42} & 2.26 &  & \textbf{0.33} & \textbf{3.42} & \textbf{2.09}\tabularnewline
\hline 
DA & Train & 0.97 & 3.20 & 1.82 &  & 0.98 & 2.21 & 1.10\tabularnewline
 & Test & 0.95 & 4.23 & 2.61 &  & \textbf{0.97} & \textbf{3.97} & \textbf{2.44}\tabularnewline
\cline{2-9} 
DA EMSC & Train & 0.97 & 2.90 & 1.64 &  & 0.98 & 2.51 & 1.30\tabularnewline
 & Test & 0.96 & 3.97 & 2.52 &  & \textbf{0.98} & \textbf{3.28} & \textbf{1.89}\tabularnewline
\cline{2-9} 
Ext. DA & Train & 0.98 & 2.15 & 1.05 &  & 0.98 & 2.33 & 1.17\tabularnewline
 & Test & \textbf{0.38} & 8.84 & 6.99 &  & 0.20 & \textbf{3.10} & \textbf{1.79}\tabularnewline
\cline{2-9} 
Ext. DA EMSC & Train & 0.98 & 2.13 & 1.03 &  & 0.98 & 2.31 & 1.16\tabularnewline
 & Test & 0.15 & 3.27 & 2.06 &  & \textbf{0.52} & \textbf{1.80} & \textbf{0.88}\tabularnewline
\hline 
\end{tabular}
\end{table*}
 The neural network models show better performance for most of the
datasets with respect to the RMSE and Huber loss, whereas the correlation
coefficients are more or less on par for the two model types. The
best performance is observed with the datasets that are subjected
to data augmentation with subsequent EMSC treatment. However, the
test set for the extrapolation datasets is not the same as for the
other datasets, and the results should not be compared directly. EMSC
in all cases improve model performance for the neural networks when
comparing the pairs: Standard - EMSC, Ext. - Ext. EMSC, DA.-DA. EMSC
and Ext. DA-Ext.DA EMSC, whereas the same does not seem to be the
case for the PLS models. Using data augmentation seem to improve the
neural network models with respect to the Huber loss, but not necessarily
the RMSE. The largest improvement when using data augmentation is
seen with the extrapolation datasets. Data augmentation seem to deteriorate
PLS model performance unless it is combined with EMSC treatment. Figure
\ref{fig:Scatter-plot}
 shows a side-by-side scatter plot comparison of the true and predicted
reference values using the PLS and Neural network model on the Extrapolation
dataset with data augmentation and subsequent EMSC treatment. There
is not much difference in the training and validation sets, whereas
the predictions are much closer to the ideal line for the neural network
models predictions. The scatter plots show clusters, which probably
corresponds to the different batches of tablets that was produced
when creating the original dataset.\cite{ShootoutWayback}

\section*{Discussion}

The two different models were not treated consistently when tuning
the two different model types. The PLS models were given an advantage
by giving the tuning loop access to information about the external
test set. This is not good modelling practice as it leads to overestimation
of the real performance. However, this was done to counteract our
biased attention to the neural network modelling. The PLS models thus
represents theoretical optimal models attainable with the choice of
preprocessing, where it is assumed that the perfect number of components
could be found. More rigorous tuning keeping the test subset completely
out of the tuning loop with either fixed validation sets or using
10 fold cross validation gave other numbers of expected optimal components
and worse results. The CNN models performed better in the benchmark
even though they were automatically tuned without any access to the
external test set from instrument two. It is of course possible that
the performance of the PLS models could be further improved by an
optimal choice of preprocessing steps, but this endeavor is best left
to a method unbiased third party evaluator.

\begin{figure*}
\caption{\label{fig:DA_EMSC_example}Example plot of data augmentation and
data augmentation with extended multiplicative scatter correction
(EMSC) of a single spectrum. A: The original spectrum is shown as
solid blue line with the data augmented spectrums as dashed lines.
B: The EMSC treatment employed nearly removes the data augmentation,
but not completely as shown on the zoomed inset.}

\includegraphics[width=1\textwidth]{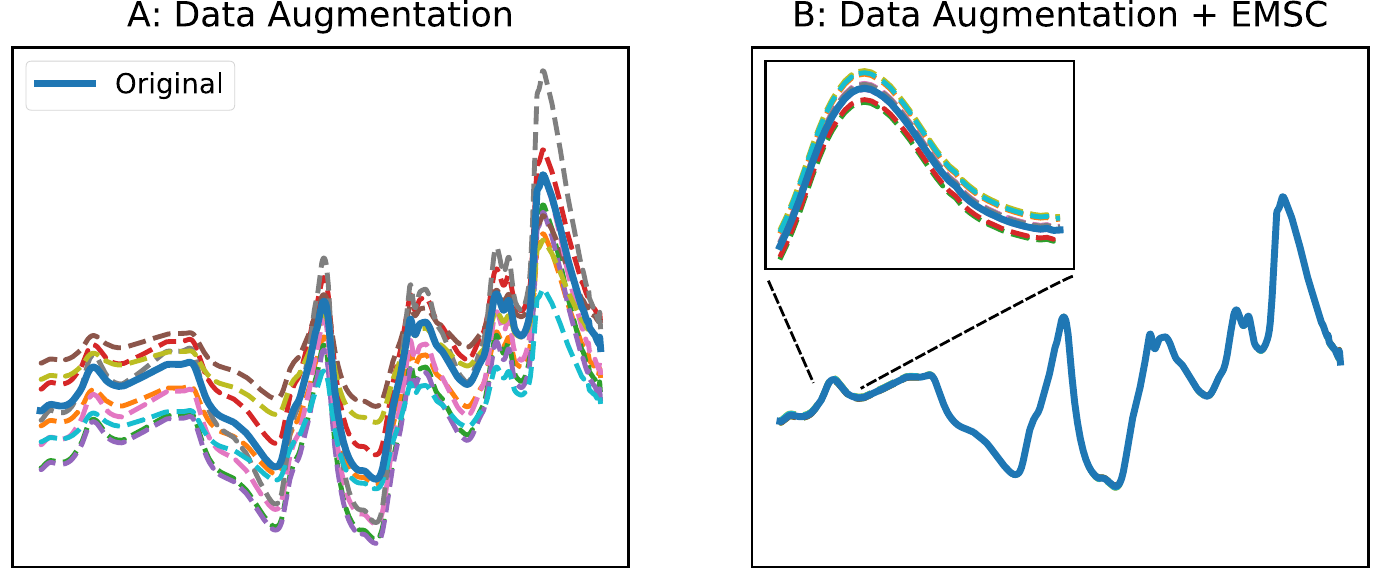}
\end{figure*}

The results showed that data augmentation improve performance when
training neural networks, which was expected as this has been shown
in other domains\cite{Bjerrum2017,Wang}. It was, however, a surprise
that the combination of data augmentation and EMSC in combination
was the best preprocessing method as the methods in theory should
counteract each other. The data augmentation tries to simulate various
scattering and offsets in the spectrum, whereas the EMSC tries to
remove these effects. Figure \ref{fig:DA_EMSC_example}
 shows an example data augmentation of a single spectrum side-by-side
with subsequent EMSC treatment. The left hand side clearly shows the
variation created by the data augmentation, which the EMSC treatment
then removes on the right-hand side, where the corrected spectrums
lie on top of each other. However, the correction is not perfect,
as visible on the zoomed inset. Residual variations are left from
data augmentation even after EMSC treatment. These residuals may help
the neural network focus its attention on the features that are invariant
to these small shortcomings of the EMSC procedure, while at the same
time benefit from the overall baseline correction from the EMSC procedure.

The extrapolation test set showed good results in the benchmark, but
only if the hyperparameters were optimized with an extrapolation validation
set and not randomly selected validation set. Using random selection
for the validation set or cross validation gave worse results in the
benchmark (results not shown). It thus seems like the network hyperparameters
need to be optimized to perform well in a task that as close as possible
mimics the wanted future performance. The extrapolation set was originally
created to test if the networks had modelled the spectra as memberships
to batches, rather than finding the inherent concentration difference
between the batches, which are visible as clusters in the plots (c.f.
Figure \ref{fig:Scatter-plot}). The CNN models could extrapolate,
which may be due to the choice of linear and semi-linear activation
functions when creating the CNN models. Extrapolation is otherwise
a feature missing from many non-linear models and in this respect,
the choice of activation functions may be crucial.

\paragraph*{Kernel activations as Spectral features}

\begin{figure}
\caption{\label{fig:The-five-most_conv1}The five most active kernels for convolutional
layer 1 for an example spectrum from the EMSC dataset (orange lines).
The activation is shown as green lines and as the blue shading in
the background.}

\includegraphics[width=1\columnwidth]{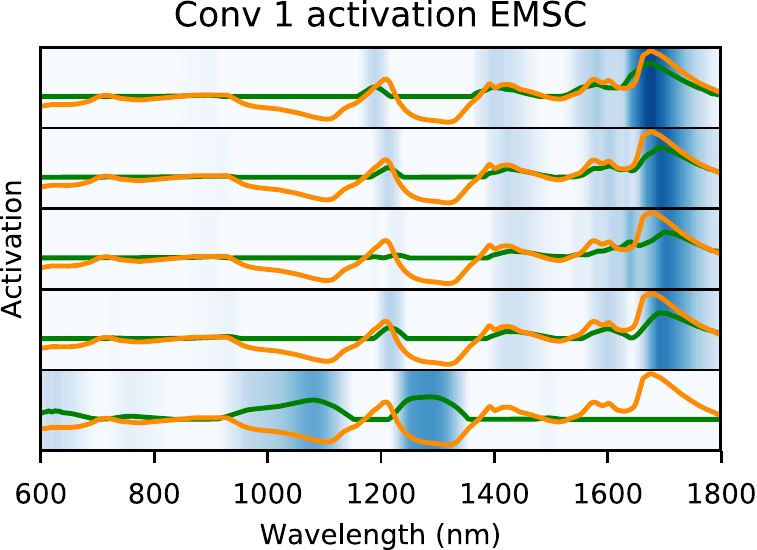}
\end{figure}

\begin{figure}
\caption{\label{fig:The-five-most_conv2}The five most active kernels for convolutional
layer 2 for an example spectrum from the EMSC dataset (orange lines).
The activation is shown as green lines and as the blue shading in
the background.}

\includegraphics[width=1\columnwidth]{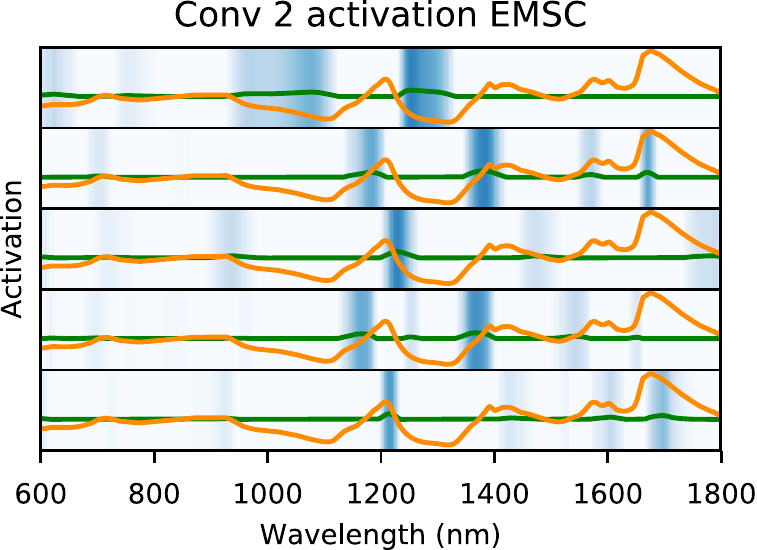}
\end{figure}

To get a better understanding of the convolutional kernels, the activation
of the kernels were plotted together with the average spectrum used
to create the activations. Examples of the five most active kernels
in convolutional layer one and two are shown in Figure \ref{fig:The-five-most_conv1}
and \ref{fig:The-five-most_conv2}, respectively.
 The plots show that the four most activated kernels in layer one
has a high activation for areas of the spectrum with high values,
whereas the lower valued areas are completely ignored. This reflects
the rectified linear unit activation function, that is zero below
a given threshold but otherwise linear. However, the activation of
the kernels also seem dependent on the slope of the spectrum rather
than the absolute value, the top points of the green lines (the activation)
are offset from the top points of the orange lines (the spectrum)
and the four first kernels differ slightly in this respect. The kernels
thus seem to combine derivative, smoothing and threshold in varying
degrees. Smoothing and using derivatives are wellknown signal processing
methods as example from the wellknown Savitsky-Golay filtering and
derivatisation.\cite{savitzky1964smoothing} The fifth kernel example
from the convolutional layer one are instead exclusively activated
at low values of the spectrum and thus more or less orthogonal to
the first ones. Using orthogonal features or components are the hallmark
of PLS and PCA modelling. The first convolutional layer thus seems
to do wellknown signal processing at least seen from a qualitative
assessment. The next convolutional layer has more complex features
(Figure \ref{fig:The-five-most_conv2}). They use convolutions from
the previous activations in both the spectral dimension but also the
kernel dimension, and can thus make non-linear combination of these
first abstractions into more complex features. The activations seen
are much more specific and narrow. There is an incomplete redundancy
between the kernels. The kernel example 2 and 4 seem similar with
respect to the two most activated areas, but have differences in other
areas. This is expected as the dropout after this layer in the neural
network encourages the network to develop redundancy during training,
as the redundant kernels can then substitute for each other, should
their activation for the next layer be dropped during training. The
kernels still seem to be activated depending on the slope of the spectrum,
but also dependent on the actual position on the spectrum. Some areas
of the spectrum which seemingly have the same slope as activated areas
are not activated at all for some of the kernels. This indicates that
the kernels are working somehow like variable selection, as they are
also dependent on the precise surroundings as seen in the spectrum.
Variable and spectral region selection is also wellknown in spectroscopic
analysis. This qualitative assessment of the kernels give an understanding
about how the network can optimize itself to the task at hand and
efficiently reinvent or mimic wellknown chemometric preprocessing
and methods. The networks are thus not completely black box predictions
as the processes up through the layers can at least be qualitatively
understood and recognized, even though the precise combinations of
smoothing, derivatisation and selection may be difficult to precisely
decipher.

While working with the models they were found to be extremely hard
to overfit, which was surprising taken the large numbers of weights
in them (several million). The convolutional layers in the beginning
on the other hand, have a very low number of weights and must work
on all the segments from the spectrum. These first layers thus seem
to work efficiently as regularization and noise reducers for the later
layers which have the majority of the connections and weights, and
thus efficiently reduce the problem of overfitting to noise in the
data. 

CNN models thus display a range of features that make them well suited
for spectroscopic analysis. They can eliminate some of the need for
preprocessing by mimicking known preprocessing steps such as smoothing,
derivatisation and region selection, they can extrapolate in assay
values and across instruments. They seem robust to overfitting due
to the regularization effects of the convolutional layers. Using convolutional
architectures for spectroscopy is thus likely to completely substitute
the dense neural network architectures which have otherwise been employed
in spectroscopy.

\section*{Conclusion}

Convolutional neural networks were used to model compound concentration
in tablets, with dropout as a regularization technique. The networks
could produce good results without preprocessing, but data augmentation
and EMSC preprocessing were both beneficial. The counterintuitive
combination of data augmentation simulating offsets in slope and intensity
and subsequent removal of the offsets with EMSC worked the best, as
there were residuals of the data augmentation even after EMSC correction.
The neural networks showed little tendency to overfitting and in nearly
all tests surpassed standard PLS models, which was otherwise hypothetical
optimal with respect to the chosen number if components. Qualitative
assessment of the trained kernel activations showed that they can
work as smoothing, derivative/slope recognisers, threshold and spectral
region selections. The CNN models were also shown to be excellent
in a hard extrapolation test, where they predicted samples from higher
assay values than available in the training set and across spectrums
recorded on different instruments.

\section*{Conflict of interests}

E. J. Bjerrum is the owner of Wildcard Pharmaceutical Consulting.
The company is usually contracted by biotechnology/pharmaceutical
companies to provide third party services. 

\bibliographystyle{elsarticle-num}
\bibliography{DeepNir_litterature}

\end{document}